\documentclass[12pt,final]{elsarticle}
\journal{}

\usepackage{graphicx,multicol,newtxtext,newtxmath,url}
\begin{document}

\begin{frontmatter}

\title{Deep Randomized Neural Networks}

\author[Pisa]{Claudio Gallicchio}
\ead{gallicch@di.unipi.it}

\author[Roma]{Simone Scardapane}
\ead{simone.scardapane@uniroma1.it}

\address[Pisa]{Department of Computer Science, University of Pisa (Italy) }
\address[Roma]{ Department of Information Engineering, Electronics and Telecommunications\\Sapienza University of Rome (Italy)}

\begin{abstract}
Randomized Neural Networks explore the behavior of neural systems where the majority of connections are fixed, either in a stochastic or a deterministic fashion. 
Typical examples of such systems consist of multi-layered neural network architectures where the connections to the hidden layer(s) are left untrained after initialization.
Limiting the training algorithms to operate on a reduced set of weights inherently characterizes the class of Randomized Neural Networks with a number of intriguing features.
Among them, the extreme efficiency of the resulting learning processes is undoubtedly a striking advantage with respect to fully trained architectures. Besides, despite the involved simplifications, randomized neural systems possess remarkable properties both in practice, achieving state-of-the-art results in multiple domains, and theoretically, allowing to analyze intrinsic properties of neural architectures (e.g. before training of the hidden layers' connections).\\
In recent years, the study of Randomized Neural Networks has been extended towards deep architectures, opening new research directions to the design of effective yet extremely efficient deep learning models in vectorial as well as in more complex data domains.
This chapter surveys all the major aspects regarding the design and analysis of Randomized Neural Networks, and some of the key results with respect to their approximation capabilities. 
In particular, we first introduce the fundamentals of randomized neural models in the context of feed-forward networks (i.e., Random Vector Functional Link and equivalent models) and convolutional filters, before moving to the case of recurrent systems (i.e., Reservoir Computing networks). For both, we focus specifically on recent results in the domain of deep randomized systems, and (for recurrent models) their application to structured domains.
\end{abstract}
\end{frontmatter}

\section{Introduction}
%: why should I care about randomization?}
\label{sec.introduction}
% why should I care about randomization?

The relentless pace of success in deep learning over the last few years has been nothing short of extraordinary. After the initial breakthroughs in the ImageNet competition \cite{lecun2015deep}, a popular viewpoint was that deep learning represented a significant shift away from hand-designing features to learning them from data. However, the majority of researchers today would agree that the shift can be more correctly classified as moving towards hand-designing \textit{architectural biases} in the networks themselves \cite{ulyanov2018deep}. This, combined with the flexibility of stochastic gradient descent and automatic differentiation, goes a long way towards explaining many of the recent advances in neural networks.

In this chapter, we consider how far we can go by relying almost exclusively on these architectural biases. In particular, we explore recent classes of deep learning models wherein the majority of connections are \textit{randomized} or more generally fixed according to some specific heuristic. In the case of shallow networks, the benefits of randomization have been explored numerous times. Among other things, we can mention the original perceptron architecture \cite{rosenblatt1958perceptron}, random vector functional-links \cite{pao1992functional,pao1994learning}, stochastic configuration networks \cite{wang2017stochastic,wang2017robust,wang2018deep}, random features for kernel approximations \cite{hamid2013compact,kar2012random,le2013fastfood,rahimi2007random,rahimi2008uniform,rahimi2009weighted}, and reservoir computing \cite{jaeger2001echo,lukovsevivcius2009reservoir}. In general, these models trade-off a (possibly negligible) part of their accuracy for training processes that can be orders of magnitude faster than fully trainable networks. In addition, randomization makes them particularly attractive from a theoretical point of view, and a vast literature exists on their approximation properties.

Differently from previous reviews \cite{scardapane2017randomness,gallicchio2017randomized,wang2016editorial}, in this chapter we focus on recent attempts at extending these ideas to the deep case, where a (possibly very large) number of hidden layers is stacked to obtain multiple intermediate representations. Extending the accuracy/efficiency trade-off also for deep architectures is not trivial, but the benefits of being able to do so are vast. As we show in this chapter, several alternatives exist for obtaining extremely fast and accurate randomized deep learning models in a variety of scenarios, especially whenever the dataset is medium or medium-to-large in size. We also comment on a number of intriguing analytical and theoretical properties arising from the study of deep randomized architectures, from their relation to kernel methods and Gaussian processes \cite{daniely2016toward}, to metric learning \cite{giryes2016deep}, pruning \cite{ramanujan2019s}, and so on. Importantly, randomization allows to potentially blend non-differentiable components in the architecture (e.g., Heaviside step functions \cite{kawaguchi2018deep}), further extending the toolkit available to deep learning practitioners.

Because we touch on a number of different fields, we do not aim at a comprehensive survey of the literature. Rather, we highlight general ideas and concepts by a careful selection of papers and results, trying to convey the widest perspective. When possible, we also highlight points that in our opinion have been under-explored in the literature, and possible open areas of research. Finally, we consider a variety of types of data, ranging from vectors to images and graph-based datasets.

\subsection*{Organization}
\label{sec:organization}
The rest of the chapter is organized in two broad parts, each further subdivided in two. We start with shallow, feedforward networks in Section \ref{sec.ff}. Because our focus is on deep models, we only provide basic concepts, and provide references and pointers to more comprehensive expositions of shallow randomized models when necessary. Building on this, Section \ref{sec.deepff} describes a selection of topics and papers pertaining to the analysis, design, and implementation of deep randomized feedforward models. Sections \ref{sec.rec} and \ref{sec.deeprc} replicate this organization for recurrent models: we first introduce the basic reservoir computing architecture in Section \ref{sec.rec} (with a focus on echo state networks), exploring their extension to multiple hidden layers and structured types of data in Section \ref{sec.deeprc}. We conclude with several remarks in Section \ref{sec.conclusions}. 

\subsection*{Notation}
%\label{sec.introduction}
%
We use boldface notation for vectors (e.g., $\mathbf{v}$) and matrices (e.g., $\mathbf{X}$). Subscripts are used to denote a specific unit inside a layer, and superscripts are used for denoting a specific layer. An index $t$ in brackets is used for time dependency. For example, $x_i^l(t)$ denotes the $i$th unit of the $l$th layer at time $t$.

\section{Randomization in Feed-forward Neural Networks}
\label{sec.ff}
As we stated in the introduction, neural networks with a single hidden layer whose connections are fixed (either randomly or otherwise) have a long history in the field, dating back to some of the original works on perceptrons. Random vector functional-links (RVFLs), originally introduced and analyzed in the nineties \cite{igelnik1995stochastic,igelnik1999ensemble,pao1992functional,pao1994learning} represent the most comprehensive formalization of this idea, with further innovations and applications up to today \cite{alhamdoosh2014fast}. In this section we provide an overview of their design and approximation capabilities, and refer to \cite{scardapane2017randomness} for a more thorough overview on their history, and to \cite{wang2017stochastic,wang2017robust,wang2018deep} for further developments in the context of these models.

\subsection{Description of the model}
\label{sec:description_of_the_model}
Consider a generic function approximation task, where we denote by $\mathbf{x}$ the input vector, by $y$ the output (e.g., a binary $\left\{0, 1\right\}$ for classification), and by $f(\mathbf{x})$ the model we would like to train. In particular, the basic RVFL model is defined as \cite{igelnik1995stochastic}:
\begin{equation}
    f(\mathbf{x}) = \sum_{i=0}^H \beta_i h_i(\mathbf{x}) \,,
    \label{eq:rvfl}
\end{equation}
where the functions $h_i(\mathbf{x})$ extract generic (fixed) features, which are linearly combined through the adaptable coefficients $\beta_i$. An example are sigmoidal basis expansions with random coefficients $\mathbf{w}_i$ and $b_i$:
\begin{equation}
    h_i(\mathbf{x}) = \frac{1}{1 + \exp\left[ -\mathbf{w}_i^T\mathbf{x} - b_i \right]} \,.
    \label{eq:sigmoid}
\end{equation}
In general, we also consider $h_0(\mathbf{x}) = 1$ to add an offset to the model, and we can also include the original input features in the output layer (similar to modern residual connections in deep networks). Assuming that the parameters in \eqref{eq:sigmoid} are all selected beforehand (e.g., by randomization), the final model in \eqref{eq:rvfl} is a linear model $f(\mathbf{x}) = \mathbf{\beta}^T h(\mathbf{x})$, where we stack in two column vectors $h(\mathbf{x})$ and $\mathbf{\beta}$ all feature expansions and output coefficients respectively. As a result, all the theory of linear regression and classification can be applied almost straightforwardly \cite{scardapane2017randomness}.

Approximation capabilities for this class of networks have been studied extensively \cite{igelnik1995stochastic,pao1994learning,gorban2015approximation,rudi2016generalization,rahimi2007random,rahimi2008uniform}. In general, RVFL networks retain the universal approximation properties of fully-trainable neural networks, with an error that decreases in the order $\frac{1}{\sqrt{B}}$. The practical success of the networks depends strongly on the selection of the random coefficients, with recent works exploring this subject at length \cite{wang2017stochastic}.

\subsection{Training the network}
\label{sec.training}
We dwell now shortly on the topic of training and optimizing standard RVFL networks. In fact, speed of training (while maintaining good nonlinear approximation capabilities) is one of the major advantages of randomized neural networks and, conversely, keeping this accuracy/efficiency trade-off is one of the major challenges in the design of deeper architectures.

Consider a dataset of desired input/output pairs $\left\{(\mathbf{x}_i, y_i)\right\}_i$. We initialize the input-to-hidden parameters randomly, and collect the corresponding feature expansions $h(\mathbf{x}_i)$ row-wise in a matrix $\mathbf{H}$. While many variants of optimization are feasible \cite{scardapane2017randomness}, by far the most common technique to train an RVFL net is to formulate the optimization problem as an $\ell_2$-regularized least squares:
\begin{equation}
    \mathbf{\beta} = \arg\min \left\{ \lVert \mathbf{H}\mathbf{\beta} - \mathbf{y} \rVert^2 + \lambda \cdot \lVert \mathbf{\beta} \rVert^2 \right\} \,,
    \label{eq:least_squares_training}
\end{equation}
where $\mathbf{y}$ is the vector of targets and $\lambda$ a free (positive) hyper-parameter. The reason \eqref{eq:least_squares_training} is a popular approach relies on (i) its strong convexity (resulting in a single minimizer), and (ii) the linearity of its gradient. The latter is especially important, since for most medium-sized datasets the problem \eqref{eq:least_squares_training} can be solved immediately as:
\begin{equation}
    \mathbf{\beta} = \left(\mathbf{H}^T\mathbf{H} + \lambda\mathbf{I}\right)^{-1}\mathbf{H}^T\mathbf{y} \,,
\end{equation}
where $\mathbf{I}$ is the identity matrix of appropriate shape, or alternatively (if $B$ is much larger than the number of points in the dataset) as:
\begin{equation}
    \mathbf{\beta} = \mathbf{H}^T\left(\mathbf{H}\mathbf{H}^T + \lambda\mathbf{I}\right)^{-1}\mathbf{y} \,.
\end{equation}
In general, solving the previous expressions has a cost which is cubic in the number of feature expansions or in the number of data points, depending on the specific formulation being chosen. For large scale problems, many ad-hoc implementations \cite{fan2008liblinear} and algorithmic advances \cite{yuan2012recent} are available to solve the problem in a fraction of the cost of a standard stochastic gradient descent. Note how, in both formulations, the term weighted by $\lambda$ acts as a numerical stabilizer on the diagonal of the matrix being inverted.

Clearly, a wide range of variants on the basic problem in \eqref{eq:least_squares_training} are possible, almost all of them loosing the possibility of a closed-form solution. Of these, we mention two that are relevant to the following. First, when considering binary classification tasks (in which the target variable is constrained as $y_i \in \left\{0,1\right\}$), we can reformulate the problem in a logistic regression fashion:
\begin{equation}
    \mathbf{\beta} = \arg\min \sum_i \big[ \mathbf{y} \odot \log\left(\sigma\left( \mathbf{H}\mathbf{\beta} \right)\right) + (1-\mathbf{y}) \odot \log\left(\sigma\left( 1 - \mathbf{H}\mathbf{\beta} \right)\right) \big] \,,
\end{equation}
where $\sigma(\cdot)$ is the sigmoid operation from \eqref{eq:sigmoid} and $\odot$ denotes elementwise multiplication. The use of the sigmoid ensures that the output of the RVFL network can be interpreted as a probability and can be used in later computations about the confidence in the prediction. Second, replacing the squared $\ell_2$ norm in \eqref{eq:least_squares_training} with the $\ell_1$ norm $\lVert \mathbf{\beta} \rVert_1$ results in sparse weight vectors, which can make the network more efficient \cite{bach2012optimization,cao2014sparse}.

\subsection{Additional considerations}
Clearly, this is only intended as a very brief introduction to the topic of (shallow) 
RVFL networks, and we refer to other reviews for a more comprehensive treatment \cite{zhang2015comprehensive,liwang2017,scardapane2017randomness,gallicchio2017randomized}. There is a pletora of interesting topics on which we skip or only brief touch, including ensembling strategies \cite{alhamdoosh2014fast} and recent works on selecting the optimal range for the pseudo-random parameters \cite{wang2017stochastic}. More generally, albeit we focus on the RVFL terminology, this class of networks has a rich history in which similar ideas have been reintroduced multiple times under different names (see also \cite{scardapane2017randomness}), so interesting pointers can be found in the literature on random kernel features \cite{rahimi2008uniform}, the no-prop training algorithm \cite{widrow2013no}, and several others. All of these works play on the delicate trade-off between keeping nonlinear approximation capabilities without sacrificing efficiency or, possibly, analytic solutions. 

We now turn to the topic of extending these capabilities to the `deep' case. Differently from the fully-trainable case, where stacking several adaptable layers can be easily justified empirically (and does not change the nature of the optimization problem), in the randomized case this is not trivial. Firstly, it is unclear whether simply stacking several randomized layers can improve accuracy at all, or even distort the original information content in the inputs. Secondly, designing other strategies going beyond the simple `stack' of layers must remain sufficiently simple and efficient to contend with fully-trainable deep learning solutions (i.e., either provide gains in accuracy or order of magnitudes in improved efficiency). In the next section, we review some significant work dealing with these two questions.

\section{Deep Random-weights Neural Networks}
\label{sec.deepff}
In this section we collect and organize a series of selected works dealing with the analysis and design of deep randomized networks. This is not built as a comprehensive survey of the state-of-the-art, but rather as a set of pointers to some of the most important ideas and results coming from the recent literature.
\subsection{Analyzing randomized deep networks through the lens of kernel methods}
To begin with, consider a generic deep randomized network $f = g \circ f_R $ defined as the composition of a \textit{representation} function $f_R$ (a stack of one or more layers with random weights), and a linear model $g$ trained on top of the representations from $f_R$ (also called later a \textit{readout}). This is a relatively straightforward extension of the previous section, where we allow the matrix $\mathbf{H}$ to be generated by more complex architectures with random weights than a single, fully-connected layer. Irrespective of the accuracy of such a model, an analysis of its theoretical properties is interesting because it corresponds to investigating the behavior of a deep network in a small subspace around its random initialization. In fact, there is a vast literature showing insightful connections of this problem with the study of kernel machines and Gaussian Processes \cite{cho2009kernel,rahimi2009weighted,mairal2014convolutional,anselmi2015deep,daniely2016toward}. A general conclusion of all these works is to show that, in the limit of infinite width, deep networks with randomized weights converge to Gaussian processes.

\cite{daniely2016toward} generalizes most of the previous results for a vast class of representation functions $f_R$, whose structure can be described by a directed acyclic graph where we associate to each node a bounded activation function, comprising most commonly used feedforward and sequential networks. They show that the \textit{skeleton} of this function (i.e., the topological structure with no knowledge of the weights) is univocally associated to a kernel function $\kappa$. The representations generated by a single realization of the skeleton, obtained by sampling the weights from a Gaussian distribution with appropriately scaled variance, are in general able to approximate the kernel itself. As a result, with high probability one can find a linear predictor $g$ able to approximate all bounded functions in the hypothesis space $\mathcal{H}$ associated to $\kappa$.

A complementary class of results, based on the novel idea of the \textit{neural tangent kernel} (NTK), can be found in \cite{yang2019scaling,arora2019exact}, allowing to extend these ideas more formally to networks with weight tying (e.g., convolutional neural networks), and to neural networks with trained weights. 

\subsection{The relation between random weights and metric learning}
Another interesting class of results is obtained by \cite{giryes2016deep} and later works, who explored the effect of the randomly initialized representation function $f_R$ on the metric space in which the data resides, exploiting tools from compressive sensing and dictionary learning. Roughly speaking, if one assumes that points in the input data corresponding to separate classes have `large' angles (compared to points in the same class), then it is possible to show that $f_R$ performs an embedding of the data in which the latter angles are shrunk more than angles corresponding to points in the same cluster. With the separation among classes increasing, the embedding obtained by a deep network makes the data easier to classify by prioritizing their angle.

Differently from works described in the previous section, these results are not viable for any deep randomized network, but only for networks with random Gaussian weights and rectified linear units (ReLU) as activation functions (or similar):
\begin{equation}
    \text{ReLU}(s) = \max\left(0, s\right) \,.
\end{equation}
The previous activation is necessary to make the network sensitive to the angles between inputs, shrinking them proportionally to their magnitude. At the same time, the analysis from \cite{giryes2016deep} has several interesting practical implications. On one side, if the assumptions on the data are correct, they allow to derive some bounds between the implicit dimension of the data and the corresponding required size of the training set (see \cite[Section V]{giryes2016deep}). More in general, even if the assumptions are not satisfied, this analysis provides a justification for the good performance of deep networks in practice, by assuming that learning the linear projection is equivalent to `choosing' a suitable angle on which to perform the shrinking across classes, instead of using the angle of their principal axis.

These results directly lead to considering this class of networks for practical learning purposes. From \cite{giryes2016deep}: ``\textit{In fact, for some applications it is possible to use networks with random weights at the first layers for separating the points with distinguishable angles, followed by trained weights at the deeper layers for separating the remaining points.}'' Extensions and variations on this core concept are considered more in-depth over the next sections.

% https://arxiv.org/pdf/1310.6343.pdf

\subsection{Deep randomized neural networks as priors}
In a very broad sense, understanding the performance of deep randomized networks in practice is akin to understanding how much the spectacular results of deep networks in several domains are due to their architectures (i.e., their architectural biases), and how much can be attributed to the specific training algorithm for selecting the weights.

One key result in this sense was developed in the work on \textit{deep image priors} \cite{ulyanov2018deep}. The paper was one of the first to show that a randomly initialized convolutional neural network (CNN) contained enough structural information to act as an efficient prior in many image processing problems. The algorithm they exploit is very simple and can be summarized in a small number of steps. First, they randomly initialize a CNN $x = f(z)$, mapping from a simple latent vector $z$ to the space of images under consideration. Given a noisy starting image $x_0$ (e.g., an image with occlusions) and a loss term $E(x, x_0)$ that depends on the specific task, the parameters of $f$ are optimized on the single image:
\begin{equation}
    f^*(z) = \arg\min E(f(z), x_0) \,.
\end{equation}
The restored image is then given by $f^*(z)$. This procedure is able to obtain state-of-the-art results on several image restoration tasks \cite{ulyanov2018deep}. In their words, ``\textit{Our results go against the common narrative that explain
the success of deep learning in image restoration to the ability to learn rather than hand-craft priors; instead, random networks are better hand-crafted priors, and learning builds on this basis.}''

Along a similar line, \cite{pons2019randomly} showed that randomly initialized CNNs on several audio classification problems performed better than some hand-crafted features, especially mel frequency cepstral coefficients (MFCCs), although they are still significantly worse than their fully trained equivalent.

\subsection{Towards practical deep randomized networks: relation with pruning}
Summarizing the discussion up to this point, we saw how deep randomized networks can be helpful for analyzing several interesting properties of deep networks. From a more practical viewpoint, fully randomized networks can be used in some specific scenarios, either as priors (due to their architectural biases), or as generic feature extractors. The question remains open, however, on whether we can exploit them also as generic learning models.

One of the first works to seriously explore this possibility was \cite{rosenfeld2019intriguing}. The authors investigated the training of a deep network wherein a large percentage of weights was kept fixed to their original values. They showed that, for modern deep CNNs, it is possible to fix up to nine tenths of the parameters and train only the remaining $10\%$, obtaining a negligible drop in accuracy in several scenarios. Apart from computational savings, this finding is interesting inasmuch it allows to describe a good portion of the neural network only with the knowledge of the specific pseudo-random number generator and its initial seed \cite{rosenfeld2019intriguing}.

This line of reasoning also connects to one of the fundamental open research questions in deep learning, pruning of architectures \cite{frankle2018lottery}. In particular, even if \textit{a posteriori} (after training), a large percentage of weights in a deep network is found to be redundant and easily removable, \textit{a priori} (before training) it is very hard to train small, compact networks. The lottery ticket hypothesis \cite{frankle2018lottery} is a recent proposal arguing that the success of most deep networks can be attributed to small subsets of weights (\textit{tickets}), and the benefit of very large networks is in having and initializing a very large number of such tickets, increasing the possibility of finding good ones. The hypothesis has generated many follow-ups (e.g., \cite{frankle2019lottery,morcos2019one}), although at the moment its relation with fully randomized networks remains under-explored (with some exceptions, e.g., \cite{ramanujan2019s}). In particular, when moving to more structured types of pruning, it is found that the lottery ticket hypothesis compares worse with respect to training from scratch smaller architectures \cite{liu2018rethinking}. ``\textit{[...] for these pruning methods, what matters more may be the obtained architecture, instead of the preserved weights, despite training the large model is needed to find that target architecture.}'' In general, this points to the fact that more work on deep randomized networks and their initialization can be beneficial also to the field of model selection and architecture search. We return on this point in one of the next sections. We refer also to \cite{zhang2019all} for similar analyses layer-wise.

\subsection{Training of deep randomized networks via stacked autoencoders}
One way to combine the advantage of randomization with a partial form of training is the use of stacked autoencoders, similar to some prior work on deep learning \cite{vincent2010stacked}. An autoencoder is a neural network with one or more hidden layers that is trained to map an input $x$ (or a corrupted version thereof) to $x$, learning a suitable intermediate representation internally. 

A general recipe to combine autoencoders with RVFLs networks is as follows \cite{cecotti2016deep}:
\begin{enumerate}
    \item Initialize a random mapping $h(\mathbf{x})$ similar to Section \ref{sec:description_of_the_model}.
    \item Train the readout to map $h(\mathbf{x})$ to the original input $\mathbf{x}$, obtaining a set of weights $\mathbf{\beta}$ (through least-squares or a sparse version of it).
    \item Use $\mathbf{\beta}^T$ as the first weight matrix of a separate deep randomized network.
    \item Repeat points (1)-(3) on the embedding generated at point (3).
\end{enumerate}

A more constructive and theoretically grounded approach to the design of deep randomized networks is described in the literature on deep stochastic configuration networks \cite{wang2018deep}. Because our focus here is on RVFL networks, we refer the interested reader to \cite{wang2018deep} and papers therein for this separate class of algorithms.

\subsection{Semi-random neural networks}

Fully-trained and randomized neural networks (what \cite{arora2019exact} calls \textit{strongly-trained} and \textit{weakly-trained} networks) are only two extremes of a relatively large continuum of models, all possessing separate trade-offs concerning accuracy, speed of training, inference, and so on. As an example of a model in the middle of this range, we describe here briefly the semi-random architecture proposed in \cite{kawaguchi2018deep}.

We replace the $i$th feature expansion in the basic RVFL model \eqref{eq:rvfl} by:
\begin{equation}
    h_i(\mathbf{x}) = \sigma_s\left(\mathbf{r}_i^T\mathbf{x}\right) \cdot \mathbf{w}_i^T\mathbf{x} \,,
\end{equation}
where $\mathbf{r}_i$ is randomly sampled, $\mathbf{w}_i$ is trainable, and the activation function $\sigma_s$ is defined for a positive hyper-parameter $s$ as:
\begin{equation}
    \sigma_s(z) = z^s \cdot H(z) \,,
\end{equation}
with $H(z)$ being the step function. For example, for $s=1$ we obtain linear semi-random features, while for $s=2$ we obtain squared semi-random features. Mimicking the matrix notation of Section \ref{sec:description_of_the_model}, the feature transformation can be written as:
\begin{equation}
    \mathbf{H} = \overbrace{\sigma_s\left(\mathbf{R}\mathbf{x}\right)}^{\text{randomized}} \odot \underbrace{\mathbf{W}\mathbf{x}}_{\text{trainable}} \,,
\end{equation}
where $\odot$ is the Hadamard (element-wise) product between matrices. While apparently counter-intuitive, this model shows a number of remarkable theoretical properties, as analyzed by \cite{kawaguchi2018deep}. Among other things, a single-hidden-layer semi-random model maintains one minimum even with a non-convex optimization problem, and its extension to more than a single hidden layer has generalization bounds that are significantly better than comparable fully-trainable networks with ReLU activation functions \cite{kawaguchi2018deep}.

Irrespective of its theoretical and practical capabilities, this model shows the power of smartly combining the two words of fully-trainable deep learning with randomized (or semi-randomized) models, which we believe heavily under-explored at the moment.

\subsection{Weight-agnostic neural networks}

% https://arxiv.org/abs/1906.04358
% Random RNN: file:///C:/Users/Simone/Downloads/guessing2001.pdf

Up to now, we considered deep networks wherein a majority of the connections are randomized. However, several of the ideas that we discussed can be extended by considering networks with \textit{fixed}, albeit not randomized, weights. In fact, as we will discuss later, in the reservoir computing field this has become a fruitful research direction. In the feedforward case, we conclude here by showing a single notable result in the case of neural architecture search (NAS), the weight-agnostic neural network \cite{gaier2019weight}.

NAS is the problem of finding an optimal architecture for a specific task. A single NAS run requires a large number of models' training, and as such, it is one of the field that could benefit the most from advancements in this sense (also from an environmental point of view \cite{strubell2019energy}). The idea of weight-agnostic network is to design a network in which all weights are initialized to the same value, and the network should be robust to this value. It allows to try a huge number of architectures extremely quickly, obtaining in some scenarios very interesting results \cite{gaier2019weight}.  ``\textit{Inspired by precocial behaviors evolved in nature, in this work, we develop neural networks with architectures that are naturally capable of performing a given task even when their weight parameters are randomly sampled}'' \cite{gaier2019weight}. Note that ideas on randomized networks in NAS also have a long history, dating back to works on recurrent neural networks \cite{schmidhuber2001evaluating}.

\subsection{Final considerations}
We conclude this general overview with a small set of final remarks and considerations. Globally, we saw that deep randomized networks have attracted a large amount of interest lately as tools for the analysis and search of deep networks, going at the hearth of a historical dichotomy between the importance of the network's architecture and the selection of its weights. Practically, several ideas and heuristics have been developed to make these randomized neural networks useful in real-world scenarios. All the ideas considered here have historical antecedents. Just to cite an example, ``\textit{It has long been known that [randomized] convolutional nets have reasonable performance on MNIST and CIFAR-10. [randomized] nets that are fully-connected instead of convolutional, can also be thought of as "multi-layer random kitchen sinks, which also have a long history''} \cite{arora2019exact}.

At the same time, we acknowledge that the performance of randomized networks have not been comparable to fully trained network on truly complex scenarios such as ImageNet. This can be due to an imperfect understanding of their behavior, or it can be a fundamental limitation of this class of models. One possibility to overcome this limit could be to combine the idea of fixing part of the network, but moving beyond pure randomization. An example of this is the PCANet \cite{chan2015pcanet}, which we have not mentioned in the main text.

While deep RVFL networks show excellent accuracy / performance trade-offs on small and medium problems, this trade-off has yet to be thoroughly analyzed for larger problems. In this line, it would be interesting to evaluate deep RVFL variations on established benchmarks such as Stanford's DAWN Deep Learning Benchmark.\footnote{\url{https://dawn.cs.stanford.edu/benchmark/}}

Finally, part of this criticism can be attributed to the lack of an established codebase for this class of models. This is also an important line of research for the immediate future.

We now turn to the topic of deep randomized \textit{recurrent} neural networks.

\section{Randomization in Dynamical Recurrent Networks}
\label{sec.rec}
\begin{figure}[tb]
\centering
\includegraphics[width = 0.6\textwidth ]{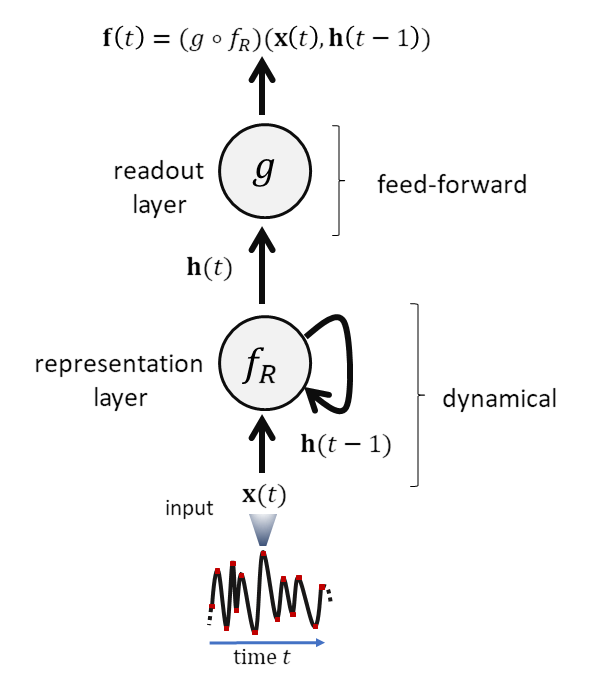}
\caption{Schematic representation of the RNN operation on temporal data.}
\label{fig.rnn}
\end{figure}

Dynamical recurrent neural models, or simply Recurrent Neural Networks \cite{kolen2001field, mandic2001recurrent}, are a widely popular paradigm for processing data that comes in the form of time-series, where each new input information is linked to the previous (and following) one by a temporal relation. Architecturally, the major difference in RNNs with respect to feed-forward neural processing systems analyzed so far is the presence of feedback among the hidden layer's \emph{recurrent} units. This is a crucial modification that makes it possible to elaborate each input in the context of its predecessors, i.e., it gives a memory to the operation of the system. Roughly speaking, apart from this architectural change, the basic description of the model does not change:
a hidden layer (made up here by recurrent units) implements a representation function $f_R$, whose outcome is tapped by a readout layer of linear units that calculate the output function $g$. 
The overall operation can be described as the composition $g \circ f_R$ (as already seen in Section
\ref{sec.deepff}).
A graphical description of this process is given in Fig.~\ref{fig.rnn}.

Going a step further into the mathematical description of the representation (hidden layer) component, we can see that its operation can be understood as that of an input-driven dynamical system. The state of such system is given by the activation of the hidden units, i.e. $\mathbf{h}(t)$. The evolution of such state is ruled by a function $f_R$ that can be formulated in several ways. For instance, in continuous-time cases such evolution function is expressed in terms of a set of differential equations, as used, e.g., in the case of spiking neural network models \cite{gerstner2002spiking}. Here, instead, we refer to the common case of discrete-time dynamical systems that evolve according to an iterated mapping of the form:
\begin{equation}
\label{eq.hidden}
\mathbf{h}(t) = f_R\big(\mathbf{x}(t), \mathbf{h}(t-1)\big) =
\sigma\big( \mathbf{W}^T \mathbf{x}(t) + \mathbf{W_R}^T \mathbf{h}(t-1)\big),
\end{equation}
where $\mathbf{W}$ and $\mathbf{W_R}$ are weight matrices that parametrize the state update function, respectively modulating the impact of the current external input and that of the previous state of the system. Typically, the activation function $\sigma$ comes in the form of a squashing non-linearity, as already examined in Section~\ref{sec.ff}.

The readout comes often in the same linear form mentioned in Section~\ref{sec.ff}, i.e., as layer of linear units that apply a linear combination of the components of the state vector: $\mathbf{\beta}^T \mathbf{h}(t)$, where the elements in  $\mathbf{\beta}$ are the parameters of the readout layer.

%UNIVERSAL APPROXIMATION.\\
Training RNN architectures implies gradient propagation across several steps: those corresponding to the length of the time-series on which the hidden layer's architecture is unrolled. It is then easy to see that training algorithms for RNN face similar difficulties to those encountered when training deep neural networks. A major related downside is that learning is computationally intensive and requires long times (an aspect partially mitigated by the availability of GPU-accelerated algorithms). As such, also in this context, the use of partially untrained RNN architectures appears immediately very intriguing. While already early works in neural networks literature pointed out the possible benefits of having untrained dynamical systems as effective neural processing models (see, e.g., \cite{albers1996dynamical}), in the last decade a paradigm called Reservoir Computing hit the literature becoming very popular as an efficient alternative to the common fully-trained design of RNNs.

\section{Reservoir Computing Neural Networks}
\label{sec.rc}
Reservoir Computing (RC) \cite{lukovsevivcius2009reservoir, verstraeten2007experimental} is a nowadays popular approach for parsimonious design of RNNs. In the same spirit of randomized neural networks approaches described in Section~\ref{sec.ff}, the basic idea of RC is to limit training to the readout part of the network, leaving the representation part unaltered after initialization. This means that the parameters (i.e., the weights) of the recurrent hidden layer are randomly initialized and then left untrained. This peculiar part of the architecture, responsible of implementing the representation function $f_R$ in Fig.~\ref{fig.rnn}, is in this context called the \emph{reservoir}.
%, as it is intended to supply a large pool of input-excitable oscillators that give a meaningful representation of the driving time-series. 
The reservoir is typically made up of a large number of non-linear neurons, and its role is essentially to provide a high-dimensional non-linear expansion of the input history into a possibly rich feature space, where the original learning problem can be more easily approached by a simple linear readout layer. 
This basic RC methodology for fast RNN set up and training has been (almost) contemporary independently proposed in literature under different names and perspectives, among which we mention Echo State Networks (ESNs) \cite{jaeger2004harnessing,jaeger2001echo}, usually with discrete-time $\tanh$ dynamics, Liquid State Machines (LSMs) \cite{maass2002real}, in the context of biologically-inspired spiking neural network models, and Fractal Prediction machines (FPMs) \cite{tino2001predicting}, originated from the study of contractive iterated function systems and fractals. Here we adopt formalism and terminology close to the prominently known ESN model.

\subsection{Reservoir Initialization}
\label{sec.rc_initialization}
Training of the readout is performed in the same way described in \ref{sec.training}, and as such we are going to discuss it further in this part.
The crucial aspect of RC networks is to guarantee a meaningful randomized initialization of the reservoir parameter, i.e., of the weight values in matrices $\mathbf{W}$ and $\mathbf{W_R}$. As we are dealing in the case with the parameters of a dynamical system, a special care needs to be devoted to the aspect of \emph{stability} of the determined dynamics. Indeed, if not properly instantiated, the reservoir system could exhibit undesired behaviors, such as instability or even chaos. If this occur, then the resulting learning model would likely respond deeply differently to very similar input time-series, thereby showing very poor generalization abilities. To account for this potential weakness, reservoirs are commonly initialized under stability properties that (in a way or another) ensure that the system dynamics will not fall into undesired regimes when put into operation. Perhaps, the most widely known of such properties is the so called \emph{Echo State Property} (ESP) \cite{yildiz2012re,jaeger2001echo,manjunath2013echo}. This is a global asymptotic (Lyapunov) stability condition on the input-driven reservoir, and essentially states that the state of the system will progressively forget its initial conditions and will depend solely on the driving input time-series. In formulas, denoting by $\tilde{f_R}(\mathbf{x_0},\mathbf{s}_N)$ the final state of the reservoir starting from initial state $\mathbf{x_0}$ and being fed by the $N$-long input time-series $\mathbf{s}_N$, the ESP can be formulated as:
\begin{equation}
\begin{array}{l}
\text{for every } \mathbf{x_0}, \mathbf{z_0}, \mathbf{s}_N:\\
\label{eq.esp}
\| \tilde{f_R}(\mathbf{x_0},\mathbf{s}_N) - \tilde{f_R}(\mathbf{z_0},\mathbf{s}_N)\| \to 0 
\text{ as } N \to \infty.
\end{array}
\end{equation}
Assuming reservoir neurons with $\tanh$ non-linearity and bounded input spaces, some baseline conditions for reservoir initialization can be derived. Specifically, a sufficient condition originates by seeing the reservoir as a contraction mapping, and requires that $\| \mathbf{W_R}\|_2 < 1$. If this condition is met, then the reservoir will show contractive behavior (and hence stability) for all possible driving inputs.
In this regard, it is worth recalling that the analysis of reservoirs as contraction mappings has also interesting connections to the resulting Markovian state space organization, the so-called \emph{architectural bias} of RNNs \cite{tino2004markovian,tivno2007markovian}. Initializing reservoirs under a contractive constraint inherently enables reservoir systems to discriminate among different input histories in a suffix-based way \cite{gallicchio2011architectural}. Interestingly, this observation explains - at least partially - the surprisingly good performance of reservoirs in many tasks (while at the same time also indicating classes of tasks that are more difficultly tackled by RC).
A necessary condition for the ESP condition assumes an autonomous reservoir (i.e., with no input) and studies its stability around the zero state. The resulting condition is given by $\rho(\mathbf{W_R}) < 1$, where $\rho(\cdot)$ denotes the spectral radius, i.e., the largest among the eigenvalues in modulus. Both the conditions are easy to implement, e.g., referring to the necessary one: after random initialization just scale the recurrent matrix by its spectral radius, and then multiply by the desired one.
Although not ensuring stability in case of non-null input, the necessary condition on the spectral radius of $\mathbf{W}_R$ is typically the one used in RC applications.

\subsection{Reservoir Richness}
\label{sec.rc_richness}
Another possible issue with untrained dynamics in RNNs is that of potential weakness of the developed temporal representations. Indeed, after contractive initialization, correlation between recurrent units activations could very high, thereby hampering the richness of the state dynamics. A simple rule of thumb here would prescribe to set the reservoir weights close to the limit of stability, e.g., by setting $\rho(\mathbf{W_R})$ to a value very close to $1$. 

Just controlling the value of spectral radius, however, could not be informative enough on the quality of the developed reservoir dynamics \cite{verstraeten2009quantification,ozturk2007analysis}. Thereby, several attempts have been done in literature to identify quality measures for reservoirs.
Notable examples are given by assessing (and trying to maximizing) information theoretic quantities, such as information storage \cite{lizier2012local}, transfer entropy \cite{schreiber2000measuring}, average state entropy \cite{ozturk2007analysis} of the reservoir over time, and entropy of individual reservoir neurons' distribution. For instance, maximizing the latter quantity led to the well-known \emph{intrinsic plasticity} (IP) \cite{triesch2005synergies,schrauwen2008improving} unsupervised adaptation training algorithms for reservoirs.

From a perspective closer to the theory of dynamical systems, several works in literature (see, e.g., \cite{legenstein2007makes, bertschinger2004real}) indicated that input-driven reservoirs that operate in a regime close to the boundary between stability and instability show higher quality dynamics. Such a region is commonly called edge of stability, edge of criticality, and also - with a slight abuse of terminology - \emph{edge of chaos}. Relevantly, reservoirs close to such a critical behavior tend to show longer short-term memory \cite{legenstein2007edge,boedecker2012information} and improved predictive quality on certain tasks \cite{livi2017determination,schrauwen2009computational,verstraeten2007experimental}. While on the one hand it could be questionable to assert that reservoirs should operate close to criticality for every learning task, on the other hand this seems a reasonable initialization condition to consider when nothing is known on the properties of the input-target relation for the task at hand. Furthermore, being able to identify the criticality would be useful to know the actual limits of reservoir stable initialization. While the identification of critical reservoir behaviour is still an open topic of research in RC community \cite{manjunath2020memory}, some (more or less) practical approaches have been introduced in literature, e.g. relying on the spectrum of local Lyapunov exponents \cite{verstraeten2009quantification}, recurrence plots \cite{bianchi2016investigating}, Fisher information \cite{livi2017determination} and visibility graphs \cite{bianchi2017multiplex}. Some works also highlighted the relation between the criticality and information theoretic measures of the reservoir \cite{boedecker2012information,torda2018evaluation}.

Another stream of RC research focuses on the idea of enforcing architectural richness in reservoir systems. Typically, reservoir units are connected by following a sparse pattern of connectivity \cite{jaeger2001echo} where, for instance, each unit is coupled only to a small constant number of others. Besides the original idea that such sparseness would have diversified the reservoir units activation (see, e.g., \cite{gallicchio2011architectural} for a counterexample), the real advantage is actually the sparsification of the involved reservoir matrices, which can sensibly cut down the computational complexity of the prediction phase. However, a related common question arising in the community is the following: \emph{is it possible to get a reservoir organization that is better than just random?}
Several literature works seem giving a positive answer to the question, pointing out approaches for effective reservoir setup. Prominent examples here are given by initialization of recurrent connections based on a ring topology \cite{rodan2011minimum, strauss2012design}, i.e., where all the units in the reservoir are simply connected to form a cycle. This kind of organization implies a number of advantages: the recurrent matrix of the reservoir is highly sparse, the stability of the system is easily controllable, the performance in many tasks is often optimized, and the resulting memorization skills are improved (approaching the theoretic limit in the linear case) \cite{tino2019dynamical,rodan2011minimum,strauss2012design}. Other common instances of constrained reservoir topologies include multi-ring reservoirs (where the recurrent neurons are connected to form more than one cycle), and chain reservoirs (where each recurrent neuron is connected only to the next one). On the one hand, these peculiar reservoir organziation can be studied from the perspective of architectural simplification \cite{strauss2012design,boedecker2009studies}, on the other hand they can find relations to the interesting concept of orthogonality in dynamical neural systems \cite{hajnal2006critical,white2004short,farkavs2016computational}. E.g., ring and multi-ring reservoirs can be seen as a very simple approach to get orthogonal recurrent weight matrices.

Another way of achieving improved quality reservoirs is to introduce depth in their architectural construction, as described in the following.

\section{Deep Reservoir Computing}
\label{sec.deeprc}
The basic idea behind the advancements on deep RNN architectures is to develop richer temporal representations that are able to exploit compositionality in time to capture the multiple levels of temporal abstractions, i.e., multiple time-scales, present in the data. This led to great success in a number of human-level applications, e.g. in the fields of speech, music and language processing \cite{hermans2013training, graves2013speech, el1996hierarchical,pascanu2013construct}. Trying to extend the randomized RC approaches described in Section~\ref{sec.rc} towards deep architectures is thereby intriguing under multiple view-points. First of all, it would enable us to analyze the bias of deep recurrent neural systems (i.e., their capabilities before training of recurrent connections). Moreover, it would make it possible to design efficient deep neural network methodologies for learning in time-series domains.

%We start our overview on deep RC from an architectural perspective.
%\subsection{Depth in Recurrent Neural Network architectures}
%\label{sec.depth}

The concept of depth in RNN design is sometimes considered questionable. Here we take a perspective similar to the authors of \cite{pascanu2013construct} and observe that even if when unrolled in time the recurrent layer's architecture becomes multi-layered, all the transitions i) from the input to the recurrent layer, ii) from the recurrent layer to the output, and iii) from the previous state to the current state \emph{are indeed shallow}. Depth can be then introduced in all of these transitions. Interestingly, some works in RC literature attempted at bridging this gap. In particular, the authors of \cite{sun2017deep} proposed a hybrid architecture where an ESN module is stacked on top of a Deep Belief Network, which introduces depth into the input-to-reservoir transition. On the other hand, the authors of \cite{bianchi2018bidirectional} proposed a RC model where a bi-directional reservoir system is tapped by a deep readout network, hence introducing depth into the reservoir-to-readout transition.
Here in the rest of this chapter we focus our analysis on the case of deep reservoir-to-reservoir transitions, where multiple reservoir layers are stacked on top of each other.
In particular, we keep our focus on the ESN formalism, extended to the multi-layer setting by the Deep Echo State Network (DeepESN) model.

\section{Deep Echo State Networks}
\begin{figure}[hph]
\centering
\includegraphics[width = 0.8\textwidth ]{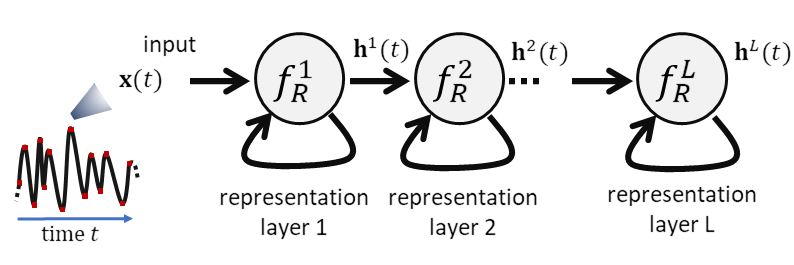}
\caption{Deep reservoir architecture.}
\label{fig.deepreservoir}
\end{figure}

DeepESNs, introduced in \cite{gallicchio2017deep}, are RC models whose operation can be described by the composition of a dynamical reservoir and of a feed-forward readout. The crucial difference with respect to standard RC is that the dynamical part is a stacked composition of multiple reservoirs, i.e., the reservoir is deep as illustrated in Fig.~\ref{fig.deepreservoir}.

The external input time-series drives the dynamics of the first reservoir in the stack, whose output then excites the dynamics of the second reservoir, and so on until the end of the pipeline. Interestingly, architecturally this corresponds to a simplification (sparsification) of a fully-connected unique reservoir (see \cite{gallicchio2017deep}).

From a mathematical perspective, the operation of the deep reservoir can be interpreted as that of a set of nested input-driven dynamical systems.
The dynamics of the first reservoir layer are ruled by:
\begin{equation}
\label{eq.layer1}
\mathbf{h}^1(t) = f_R^1\big(\mathbf{x}(t), \mathbf{h}^1(t-1)\big) = 
\sigma\big(\mathbf{W}^{1\,T} \mathbf{x}(t) + \mathbf{W_R}^{1\,T} \mathbf{h}^1(t-1)\big).
\end{equation}
While the evolution of the temporal representations developed in successive layers $l>1$ is given by:
\begin{equation}
\label{eq.layerl}
\mathbf{h}^l(t) = f_R^l\big(\mathbf{h}^{l-1}(t), 
\mathbf{h}^l(t-1)\big) = 
\sigma\big(\mathbf{W}^{l\,T} \mathbf{h}^{l-1}(t) + \mathbf{W_R}^{l \,T} \mathbf{h}^l(t-1)\big),
\end{equation}
where $\mathbf{W}^l$ denotes the weight matrix for the connections between layer $l-1$ and $l$, and $\mathbf{W_R}^l$ is the recurrent weight matrix for layer $l$. Given such a mathematical formulation, it is possible to derive stability conditions for the ESP of deep RC models. This was achieved in \cite{gallicchio2017echo} for a more general case of reservoir computing models with leaky integrator units. For the case of standard tanh neurons considered here, the sufficient condition is given by:
\begin{equation}
\label{eq.sufficient}
\max_{k=1,\ldots,L}\; \sum_{i = 1}^k\|\mathbf{W_R}^i\|_2 \prod_{j=i+1}^{k}\|\mathbf{W}^j\|_2 < 1,
\end{equation}
while the necessary one reads as follows:
\begin{equation}
\label{eq.necessary}
\max_{k=1,\ldots,L}\; \rho(\mathbf{W_R}^k) < 1.
\end{equation}
Notice that both conditions generalize (for multi-layered settings) the respective ones for shallow reservoir systems already discussed in Section~\ref{sec.rc}.

\begin{figure}[tb]
\centering
\includegraphics[width = 0.6\textwidth ]{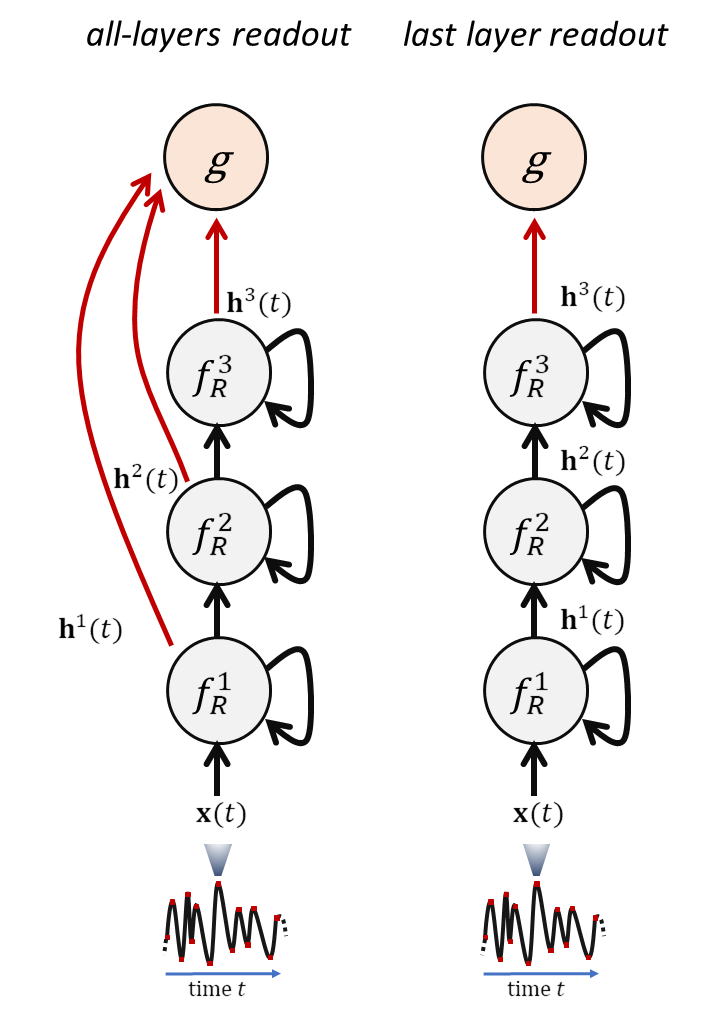}
\caption{Readout settings for DeepESN. Trained connections are only those pointing to the readout (in red).}
\label{fig.readouts}
\end{figure}

As illustrated in Fig.~\ref{fig.readouts}, there are two basic settings for the readout computation. In a \emph{all-layers} setup, the readout is fed by the activation of all the reservoir layers. In a \emph{last-layer} setup, the readout receives only the activations of the last layer in the stack. In the former case the learner is able to exploit the qualitatively different dynamics developed in the different layers of the recurrent architectures (possibly weighting them in a suitable way for the learning task at hand). In the latter case, the idea is that the stack of reservoirs has enriched the developed representations of the driving input in such a way that the readout operation can now be more effective. 
Again, training is limited to the connections pointing to the readout, and is performed as discussed in Section~\ref{sec.training}.

Interestingly, the structure that is imposed to the organization of the recurrent units in the reservoir is reflected by a corresponding structure of the developed temporal representation.
This has been analyzed recently under several points of view, delineating a pool of potential advantages of deep recurrent architectures that are independent of the training algorithms and shedding light on the architectural bias of deep RNNs.

\subsection{Enriched Deep Representations}
A first inherent benefit of depth in RNNs is given by the possibility to develop progressively more abstract representations of the driving input. In the temporal domain this means that different layers are able to focus on different time-scales, and the networks as a whole is capable of representing temporal information at multiple time-scales. A first evidence in this sense was given in \cite{gallicchio2017deep}, where it was shown that effects of input perturbations last longer in the  higher layers of the deep reservoir architecture. This important observation was in line with what reported in \cite{hermans2013training} for fully trained deep RNNs, and pointed out the great role played by the layering architectural factor in the emergence of multiple-times scales.

A further evidence of multiple-view representations in untrained RNN systems was given in \cite{gallicchio2018design}, where 
%- given a superimposition of sinusoidal functions in input - 
it was shown that the different layers in the deep reservoirs tend to develop different frequency responses (as emerging through a fast Fourier transform of the reservoir activations). 
%In particular, in the experiments shown in \cite{gallicchio2017hierarchical}, higher reservoir layers showed a progressive focus on lower frequencies in the input signal. 
This insights was  exploited  to develop an automatic algorithm for the setup of the depth in untrained deep RNN. The basic idea was to analyzing the behavior of each new reservoir layer in the architecture as a filter, stopping adding new layers when the filtering effect becomes negligible. The resulting approach, in conjunction with IP unsupervised adaptation of reservoirs, was shown to be extremely effective in speech and music processing, achieving state-of-the-art results and beating the accuracy of more complex fully-trained gated RNN architectures, requiring only a fraction of their respective training times \cite{gallicchio2018design,deepcomparison}.

Richness of deep reservoir dynamics was also explored in the context of stability of dynamical systems and local Lyapunov exponents. In this regard, the major achievement is reported in \cite{gallicchio2018local} where it was shown, both mathematically and experimentally, that organizing the same number of recurrent units into layers naturally (i.e., under easy conditions) has the effect of pushing the resulting system dynamics closer to the edge of criticality. Under a related view-point, deep RC settings were found to boost the short-term memory capacity in comparison to equivalent shallow architectures \cite{gallicchio2017deep}.
%, in both all-layers \cite{gallicchio2017deep} and last-layer \cite{gallicchio2018memory} readout settings.

More recent works on deep RC highlighted even further the role of certain aspects of network's architectural construction in the enrichment of developed dynamics. In this concern, results in \cite{gallicchio2019richness} pointed out the relevance of a proper scaling of inter-layer connections, i.e., of the weights in matrices $\mathbf{W}^l$, for $l>1$, in \eqref{eq.layerl}. It was found that such scaling has a profound impact on the quality of dynamics in higher layers of the network, with larger (resp. smaller) values leading to higher (resp. smaller) average state entropy and number of linearly-uncoupled dynamics. The importance of inter-reservoirs connectivity patterns was also pointed out in the context of spiking neural networks in \cite{zajzon2018transferring}
%Finally, the work in \cite{gallicchio2019reservoir} explored the performance of deep reservoir models with constrained recurrent topology in each layer, showing that the synergy between structuring the recurrent connectivity at both higher and lower levels can determine a decisive boost in the performance of RC.

\subsection{Deep Reservoirs for Structures}
In many real-world domains the information under consideration presents forms of aggregation that can be naturally represented by complex forms of data structures, such as trees or graphs. Learning in such structured domains opens entire worlds of application opportunities and at the same time it implies a large number of difficulties. The interested reader is referred to \cite{bacciu2019gentle} for a gentle introduction to the research field.

Here we briefly summarize the extension of RC models for dealing with trees and graphs. Starting with tree domains, the basic idea is inspired by the original concept of \emph{Recursive} Neural Networks (RecNNs) \cite{sperduti1997supervised,frasconi1998general}, and consists in applying a reservoir system to each node in the input tree, starting from the leaves and ending up in the root. 
The overall process is again seen as a composition of a representation component followed by a readout layer. In this case, the representation component is implemented by the reservoir as a state transition system that operates on discrete tree structures. 
The nodes in the input tree take the role of time-steps in the computation of conventional reservoirs, and the states of \emph{children} nodes takes the role of the previous state. With these concepts in mind, the state (or neural \emph{embedding}) computed for each node $n$ at layer $l$ can be expressed as:
\begin{equation}
\label{eq.tree}
\begin{array}{ll}
\mathbf{h}^l(n) & =  f_R \big(\mathbf{x}^l(n), \mathbf{h}^l(ch_1(n)),\ldots,\mathbf{h}^l(ch_k(n))\big) \\
\\
& = \sigma\big(\mathbf{W}^{l\,T} \mathbf{x}^l(n) + \sum_{i = 1}^k \mathbf{W_R}^{l\,T} \mathbf{h}^l(ch_i(n))\big),
\end{array}
\end{equation}
where $\mathbf{h}^l(ch_i(n))$ is the state computed by layer $l$ for the i-th child of node $n$. Note that $\mathbf{x}^l(n)$ is the input information that drives the state update at the current layer: the (external) input label attached to node $n$ for the first layer, and the state for node $n$ already computed at the previous layer, for layers  $l > 1$.

For the case of graphs the reservoir operation is further generalized, and the embedding computed for each vertex in the input structure becomes a function of the embedding developed for its \emph{neighbors}. The state transition of a deep graph reservoir system operating on a vertex $v$ at layer $l$ can be formulated as follows:
\begin{equation}
\label{eq.graph}
\begin{array}{ll}
\mathbf{h}^l(v) & =  f_R \big(\mathbf{x}^l(v),\{\mathbf{h}^l(v')\}_{v' \in \mathcal{N}(v)}\big) \\
\\
& = \sigma\big(\mathbf{W}^{l\,T} \mathbf{x}^l(v) + \sum_{v' \in \mathcal{N}(v)} \mathbf{W_R}^{l\,T} \mathbf{h}^l(v')\big),
\end{array}
\end{equation}
where $\mathcal{N}(v)$ is the neighborhood of $v$ and, as before, $\mathbf{x}^l(v)$ is the driving input information for vertex $v$ at layer $l$.

The two deep reservoir models expressed by \eqref{eq.tree} and \eqref{eq.graph} are based on randomization as conventional RC approaches, and are formalized respectively in \cite{gallicchio2019deep} and \cite{gallicchio2020fast}. Experimental assessment in these papers indicate the great potentiality of the randomization approach also in dealing with complex data structures, often establishing new state-of-the-art accuracy results on problems in the areas of document processing,  cheminformatics and social network analysis.

\section{Conclusions}
\label{sec.conclusions}
In the face of huge computational power and strong automatic differentiation capabilities exhibited by most computers and frameworks today, a focus on randomization as a quick alternative to full optimization can seem counter-productive. Yet, in this chapter we hope to have provided sufficient evidence that, despite the breakthroughs of fully-trained deep learning, randomized neural networks remain an area of research with strong promises. From a practical perspective, they can achieve significant accuracy / efficiency trade-offs in most problems, albeit strong performance on very large-scale problems currently remain difficult. From a theoretical perspective, they are an irreplaceable tool for the analysis of the properties and dynamics of classical neural networks. More importantly, we believe fully-trainable and fully-randomized networks stand at two extremes of a wide range of interesting architectures, a continuum that only today starts to be more thoroughly explored. We believe our exposition can summarize some of the most promising lines of research and provide a good entry point in this ever growing body of literature.

%\begin{acknowledgement}
%Text...
%\end{acknowledgement}
%
%\section*{Appendix}
%\addcontentsline{toc}{section}{Appendix}
%
%Text...
%
\bibliographystyle{elsarticle-num}
\bibliography{main.bib}
\end{document}